\documentclass[letterpaper]{article}
\usepackage{aaai}
\usepackage{times}
\usepackage{helvet}
\usepackage{courier}
\usepackage{cite}
\usepackage{natbib}
\usepackage{multirow}
\usepackage{graphicx}
\usepackage{pgfplots}
\usepackage{hyperref}
\usepackage{changes}
\usepackage{amssymb}
\usepackage{rotating}
\graphicspath{ {./images/} }
\title{Makeup216: Logo Recognition with Adversarial Attention Representations}

\author{
Junjun Hu\textsuperscript{*1} Yanhao Zhu\textsuperscript{1} Bo Zhao\textsuperscript{1} Jiexin Zheng\textsuperscript{1,2} Chenxu Zhao\textsuperscript{1} Xiangyu Zhu\textsuperscript{3} Kangle Wu\textsuperscript{1} Darun Tang\textsuperscript{1} \\
\textsuperscript{1} Mininglamp Academy of Sciences, Mininglamp Technology \\
\textsuperscript{2} Chang'an University \\
\textsuperscript{3} Institute of Automation, Chinese Academy of Sciences 

}

\begin{document}

\maketitle
\begin{abstract}
\begin{quote}
One of the challenges of logo recognition lies in the diversity of forms, such as symbols, texts or a combination of both; further, logos tend to be extremely concise in design while similar in appearance, suggesting the difficulty of learning discriminative representations. To investigate the variety and representation of logo, we introduced Makeup216, the largest and most complex logo dataset in the field of makeup, captured from the real world. It comprises of 216 logos and 157 brands, including 10,019 images and 37,018 annotated logo objects. In addition, we found that the marginal background around the pure logo can provide a important context information and proposed an adversarial attention representation framework (AAR) to attend on the logo subject and auxiliary marginal background separately, which can be combined for better representation. Our proposed framework achieved competitive results on Makeup216 and another large-scale open logo dataset, which could provide fresh thinking for logo recognition. The dataset of Makeup216 and the code of the proposed framework will be released soon.

\end{quote}
\end{abstract}

\section{Introduction}
\begin{figure}
\centering
\includegraphics[width=0.45\textwidth ]{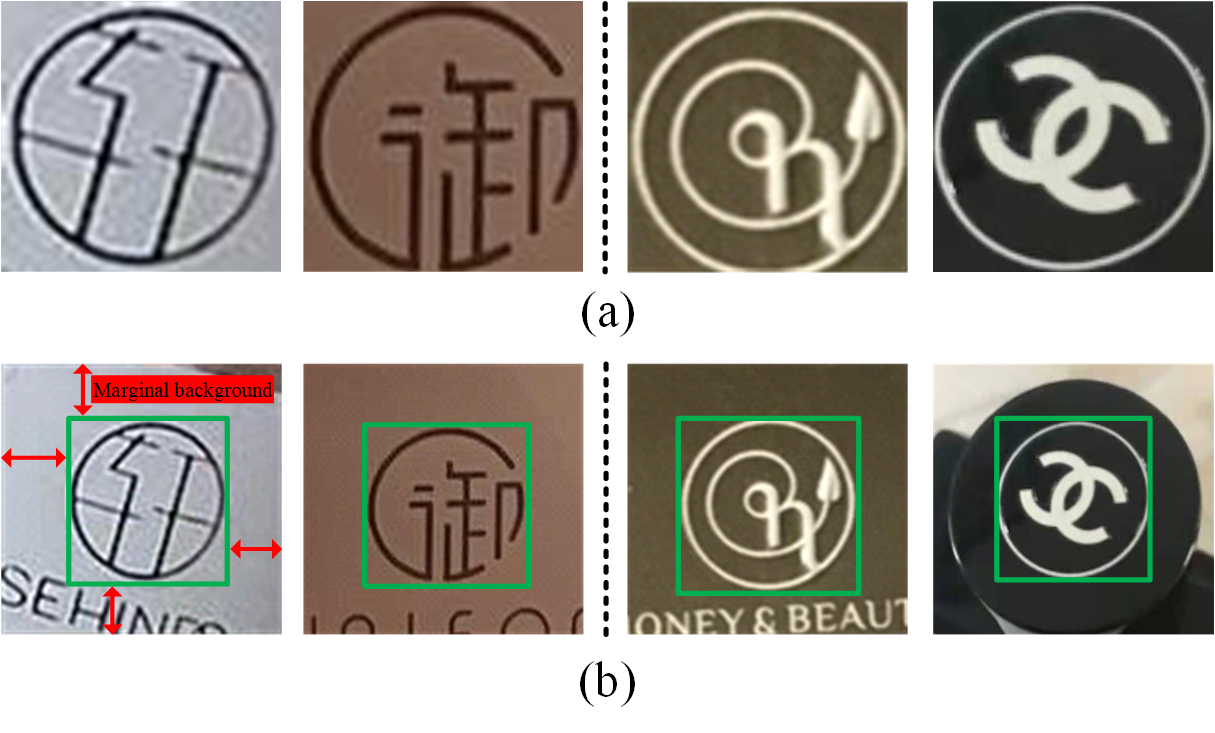}
\caption{(a) False retrieval due to the low inter-class variability of pure logos. (b) To break the limitation of pure logos, we enlarged the bounding boxes with more marginal background exposed, broader vision and context can be acquired.}
\label{figure:1}
\end{figure}

\noindent 
Logo is one of the core components of brand image, and identifying logos can statistically analyze the exposure and influence of a brand from an amount number of images/videos (\citeauthor{gao2014brand} \citeyear{gao2014brand}). Due to the design characteristics of the logos: favoring concise symbols or texts (\citeauthor{ad2012logo}\citeyear{ad2012logo}), logos do not contain rich information to be represented; further, different logos may only differ by a character or a stroke, both of which makes it very difficult to learn deep features with discrimination.

Logo recognition usually consists of two steps, locating the bounding box, and identifying the logo. Traditional logo recognition uses a close-set approach, and popular frameworks are Faster R-CNN (\citeauthor{ren2015faster} \citeyear{ren2015faster}), SSD (\citeauthor{liu2016ssd} \citeyear{liu2016ssd}), and YOLOv3 (\citeauthor{redmon2018yolov3} \citeyear{redmon2018yolov3}) etc. However, close-set is more demanding on data, and it cannot identify logos that have not appeared in the training set. To tackle this problem, open-set methods were proposed (\citeauthor{tuzko2017open} \citeyear{tuzko2017open}; \citeauthor{bastan2019large} \citeyear{bastan2019large}; \citeauthor{fehervari2019scalable} \citeyear{fehervari2019scalable}), which consists of two stages: detecting logo box by class-agnostic binary detector and identifying the detected box by classifier or retrieval. The binary-detection stage can directly use the output of the detection framework, such as Faster R-CNN and YOLOv3. The identifying stage can apply metric learning to measure the distance of nonlinear data, thus allowing complex data to be easily compared in the embedding space.  

In real application, we found that if a low threshold is set, the logo-like region can be easily detected by the class-agnostic binary detector, but it is very difficult to identify whether it is a logo and which class this logo belongs to, i.e., to map the logos well in the embedding space with similar classes clustered and different classes separated. Therefore, we believe that the main challenge lies in the second stage of logo recognition, which is our principal concern in this work.

The shortcomings of the current public logo datasets (Tab. \ref{table:1}) are: 1) a small number of logo classes, can hardly reflect practical logo recognition challenges (\citeauthor{romberg2011scalable} \citeyear{romberg2011scalable}; \citeauthor{neumann2002integration} \citeyear{neumann2002integration}); 2) crawled from the web, logos are very dominant in the images and the scene complexity are reduced (\citeauthor{wang2020logodet} \citeyear{wang2020logodet}; \citeauthor{su2018open} \citeyear{su2018open}; \citeauthor{tuzko2017open} \citeyear{tuzko2017open}); 3) image level annotation (\citeauthor{su2018scalable} \citeyear{su2018scalable}), which can not be used for logo recognition tasks. Therefore, we introduced Makeup216, of which the advantages are: 1) it is focused on the field of makeup, requiring stronger recognition algorithm to distinguish them because of the small difference between each logo; 2) it is taken from the real world, where logos all exists on real products rather than posters, a variety of complex difficulties arises, such as: blur, occlusion, mirror reflection, non-rigid deformation, etc.; 3) the object number per images is the largest among all the datasets. The advantages of Makeup216 can largely help to extend the feasibility of logo recognition model. 

We found that pure logos had a narrow field of view and limited information for recognition, which can easily lead to a false retrieval, as shown in Fig. \ref{figure:1}(a), where the top-1 recalled logos are not correct but retain large similarity. However, if more external regions are considered, we can acquire a broader vision and context information. The products behind logos and the texts/symbols adjacent to logos can help to verify the identification and avoid false recall, as shown in Fig. \ref{figure:1}(b). Based on this observation, we propose a novel logo recognition framework with an adversarial attention representation (AAR) module to make use of the marginal background. The framework can clarify what its subject is and also maintain the auxiliary information to judge more accurately. It is found that the problems in Fig. \ref{figure:1}(a) can be solved in our proposed framework, which will be elaborated in the experimental section. 

To test the noise resistance of logo recognition model, we built a distractor set for Makeup216 inspired by the image retrieval dataset (\citeauthor{radenovic2018revisiting}\citeyear{radenovic2018revisiting}); and therefore constructed two evaluation protocols, one is without distractor samples, the other is with distractor samples. The proposed framework was proved to be have greater superiority in face of distractor samples.

The main contributions of this work are as follows:
\begin{enumerate}
\item A dataset majoring in the field of makeup is introduced. The diversity of logos and complexity of scenes make it a challenging dataset for logo recognition; a novel set of evaluation protocols is also constructed which could reflect the robust of the logo recognition model. Both the dataset Makeup216 and the protocols will be open access. 
\item An adversarial attention representation (AAR) framework for logo recognition is proposed. The framework takes full advantage of the marginal auxiliary information to complement the representation of pure logos. The AAR can tell what should be focused on and what can be used as a supplement to better accuracy.
\item Our framework provides a baseline for logo recognition, and sufficient experiments on Makeup216 and LogoDet-3k indicate that it achieved a state of the art performance.
\end{enumerate}


\section{Related Work}
\begin{table}[h]
\centering

\resizebox{0.475\textwidth}{!}{
\begin{tabular}{c|c|c|c}
\hline
 Datasets& Logos & Images &Objects per image\\
 \hline
 FlickrLogos-32 (\citeauthor{romberg2011scalable} \citeyear{romberg2011scalable})& 32    &8,240&0.685 \\
 BelgaLogos (\citeauthor{neumann2002integration} \citeyear{neumann2002integration})& 37    &10,000&0.277\\
 QMUL-OpenLogo (\citeauthor{su2018open} \citeyear{su2018open})& 352    &27,083 &1.891\\
 Logos-in-the-Wild (\citeauthor{tuzko2017open} \citeyear{tuzko2017open}) & 871    &11,054&2.972\\
 LogoDet-3K (\citeauthor{wang2020logodet} \citeyear{wang2020logodet})& 3000    &158,652&1.224\\
 WebLogo-2M (\citeauthor{su2018scalable} \citeyear{su2018scalable})& 194 & 1,867,177&NULL \\
 \hline
 \textbf{Makeup216} & 216&10,019 & \textbf{3.695}  \\
 \hline
 
\end{tabular}
}
\caption{Statistical comparison of logo datasets.}
\label{table:1}
\end{table}

\subsection{Logo dataset}
Logo datasets of different sizes have been released shown in Tab. \ref{table:1}. FlickrLogos-32 (\citeauthor{romberg2011scalable} \citeyear{romberg2011scalable}), one of the small-sized datasets, has 32 logos. Medium-sized datasets, such as Logos-in-the-Wild (\citeauthor{tuzko2017open} \citeyear{tuzko2017open}) constituted of 871 logos. LogoDet-3K (\citeauthor{wang2020logodet} \citeyear{wang2020logodet}) is the largest dataset, containing 3000 logos. All of the above datasets were collected from web, it is very convenient and can cover large-scale logo classes but still has some drawbacks, for instance, some images are pure logos where the logos dominate the images without noisy backgrounds that interfere with recognition, some images are posters with a very formal presentation of the logos without complexity.

\subsection{Logo recognition}
Frameworks such as Faster R-CNN (\citeauthor{ren2015faster} \citeyear{ren2015faster}), SSD (\citeauthor{liu2016ssd} \citeyear{liu2016ssd}), and YOLOv3 (\citeauthor{redmon2018yolov3} \citeyear{redmon2018yolov3}) are widely used for logo recognition. Logo-Yolo (\citeauthor{wang2020logodet} \citeyear{wang2020logodet}) optimized anchor and loss based on YOLOv3 according to the characteristic of logo dataset. Recently, open-set methods have received more attention with the advantage of being able to predict unknown classes. They first detected logo-like boxes by a binary detector, and then extracted deep features of the logo-like boxes for retrieval. Some of the deep feature extractors are trained by classification (\citeauthor{bianco2015logo} \citeyear{bianco2015logo}; \citeauthor{hoi2015logo} \citeyear{hoi2015logo}), others are by deep metric learning (\citeauthor{tuzko2017open} \citeyear{tuzko2017open}; \citeauthor{bastan2019large} \citeyear{bastan2019large}; \citeauthor{fehervari2019scalable} \citeyear{fehervari2019scalable}). A lot of loss functions have been proposed to boost the discriminative power in the embedding space, such as Contrastive loss (\citeauthor{hadsell2006dimensionality} \citeyear{hadsell2006dimensionality}), Triplet loss (\citeauthor{schroff2015facenet} \citeyear{schroff2015facenet}), and ArcFace loss (\citeauthor{deng2019arcface} \citeyear{deng2019arcface}). 

\subsection{Representation learning}
Currently, unsupervised constrastive learning methods which can extract the inherent representation of an image in the absence of artificially annotation is emerging. Negative samples were first regarded as necessary for contrastive learning, for instance, MOCO (\citeauthor{he2020momentum} \citeyear{he2020momentum}) designed a memory encoder and maintained a negative queue to compare with positive samples. SimClR (\citeauthor{chen2020simple} \citeyear{chen2020simple}) compared augmented positive pairs and negative pairs in each batch which is very simple but requires large batches. It is very soon that BYOL (\citeauthor{grill2020bootstrap} \citeyear{grill2020bootstrap}) and SimSiam (\citeauthor{chen2021exploring} \citeyear{chen2021exploring}) were proposed where negative pairs and large batches are not necessary.  In this work, we would apply the core idea of contrastive learning without negative samples, but the difference is that one of the two inputs are not augmented view of original images but enlarged one with more marginal background exposed. 

\subsection{Adversarial learning}
 Adversarial training is defined as training on adversarial examples, and many methods (\citeauthor{shafahi2019adversarial} \citeyear{shafahi2019adversarial}; \citeauthor{zhu2019freelb} \citeyear{zhu2019freelb}) have been proposed to generate adversarial examples to improve the robustness of neural networks. Some work generated adversarial attacks for unlabeled data and combined it with contrastive learning (\citeauthor{kim2020adversarial} \citeyear{kim2020adversarial}; \citeauthor{kim2021contrastive} \citeyear{kim2021contrastive}; \citeauthor{jiang2020robust} \citeyear{jiang2020robust}). Ganin added an adversarial domain classifier in addition to the label classifier to address domain adaptation problem (\citeauthor{ganin2016domain} \citeyear{ganin2016domain}). Inspired by this idea of two different classifiers, we designed an adversarial attention representation module including positive and negative contrast sub-modules. Our aim is that the two sub-modules can learn common and uncommon features separately, which, when combined, can have a more comprehensive representation of the logos.

\section{Dataset}
\subsection{Makeup216}

\begin{figure}[h]
\centering
\includegraphics[width=0.45\textwidth ]{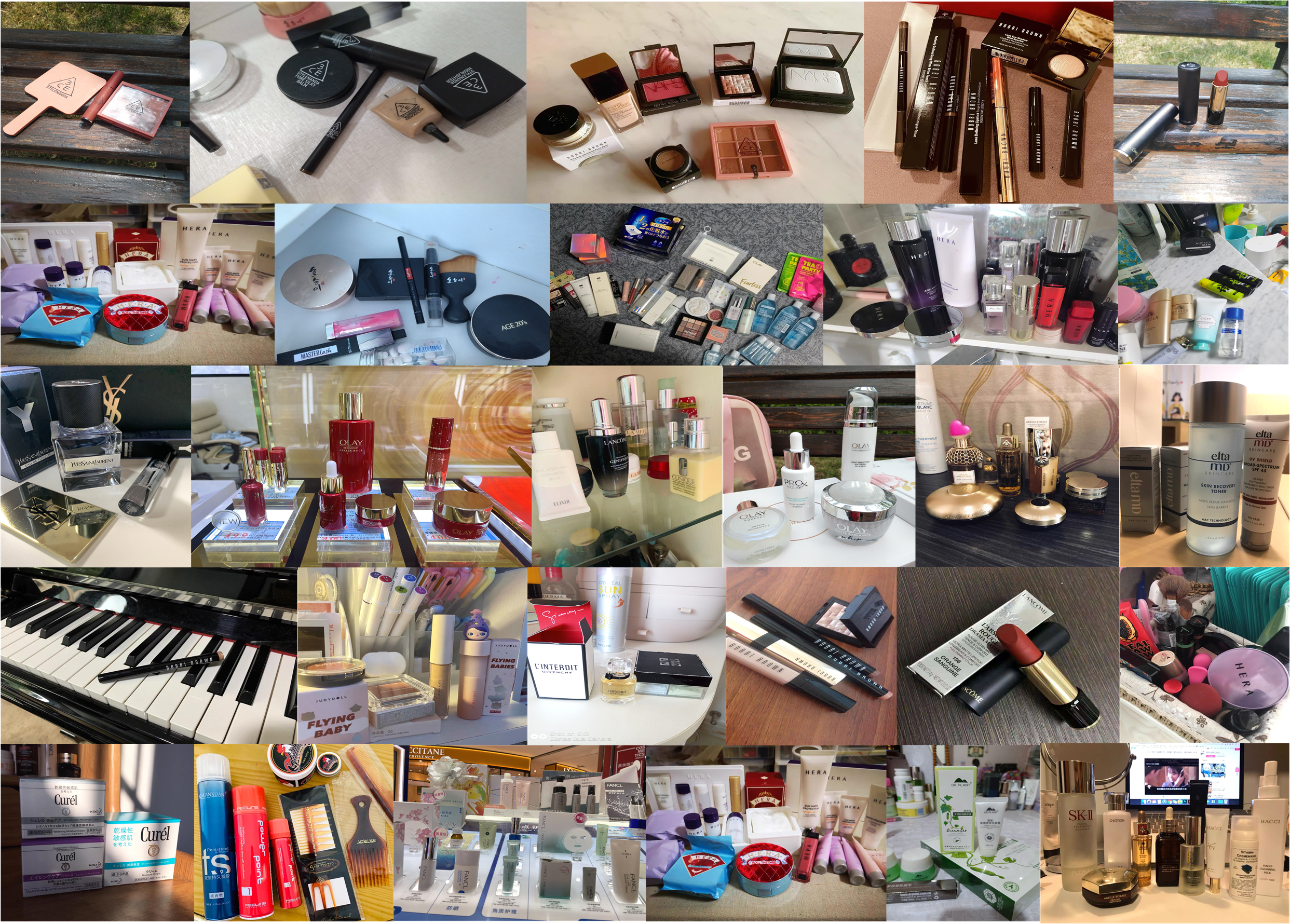}
\caption{Samples showing diversity and complexity}
\label{figure:2}
\end{figure}

To overcome the disadvantage of web crawling, we use crowdsourcing to collect logo images from the real world. The collection and pre-processing stages are described detailed in \textbf{\textit{Appendix A}}. To illustrate the diversity and complexity, we sampled typical images from Makeup216 shown in Fig. \ref{figure:2}. 


\subsubsection{Annotation}
\begin{figure}[h]
\centering
\includegraphics[width=0.40\textwidth ]{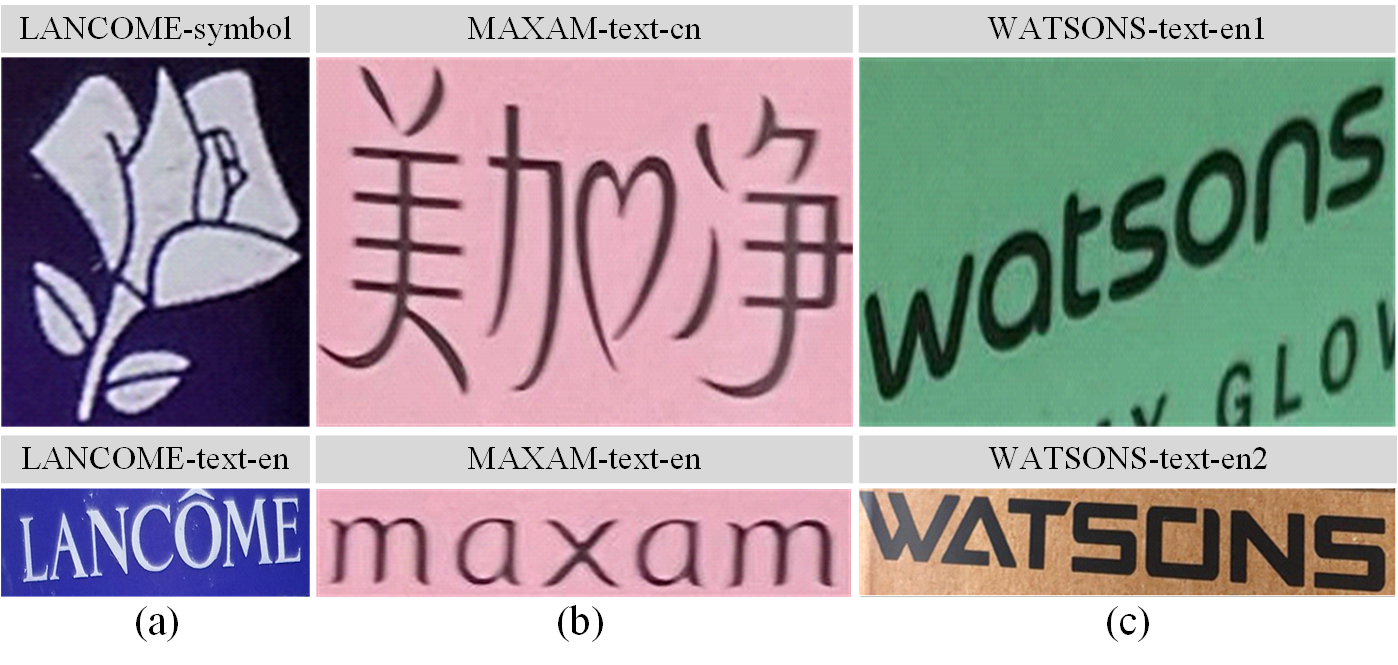}
\caption{ Annotations corresponds to multiple forms of logos: (a) Symbol and text. (b) English and Chinese texts.   (c) Two types of fonts.}
\label{figure:3}
\end{figure}

Since a brand usually owns multiple forms of logos, we added suffixes after brand as the labels. For example, the brand \textit{LANCOME} has both symbol and text logos, so the labels are annotated as \textit{LANCOME-symbol} and\textit{ LANCOME-text-en}; the language and the font of the texts are also various (Fig. \ref{figure:3}). In shot, the labels for all the logos are basically designed by the template of \textit{brand-symbol/text-language}, and suffix number is added if more than one types of fonts exist.

\subsubsection{Statistics}
\begin{figure}[h]
\centering
\includegraphics[width=0.45\textwidth ]{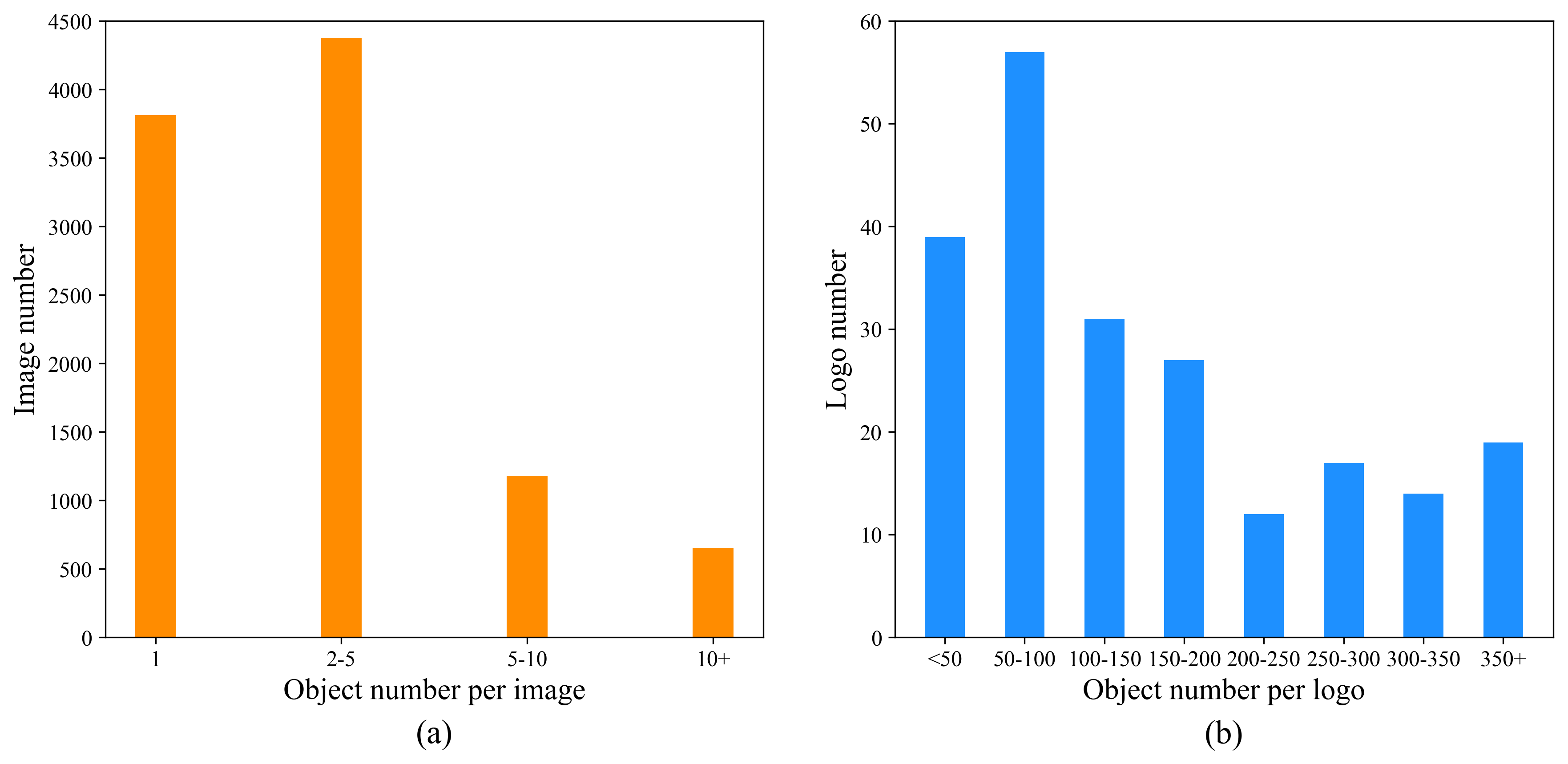}
\caption{ (a) Distribution of the number of logo objects per image. (b) Distribution of the number of objects per logo. }
\label{figure:4}
\end{figure}

Makeup216 consists of 216 logos and 157 brands, with 10,019 images and 37,018 annotated logo objects. The distribution of the number of logo objects per image is shown in Fig. \ref{figure:4}(a). It can be seen that most of the images contain more than one object, with the peak number ranging from 2 to 5. The number of objects per logo is different, and the peak number is between 50 and 100, with the majority being below 200 (Fig. \ref{figure:4}(b)). We split trainval and test sets in the ratio of 8:2, and then split trainval set into a training set and a validation set by the ratio of 8:2 as well. Detailed image number of training, validation and test sets is 6,307, 1,647, 2,065.

\begin{figure*}[h]
\centering
\includegraphics[width=0.9\textwidth ]{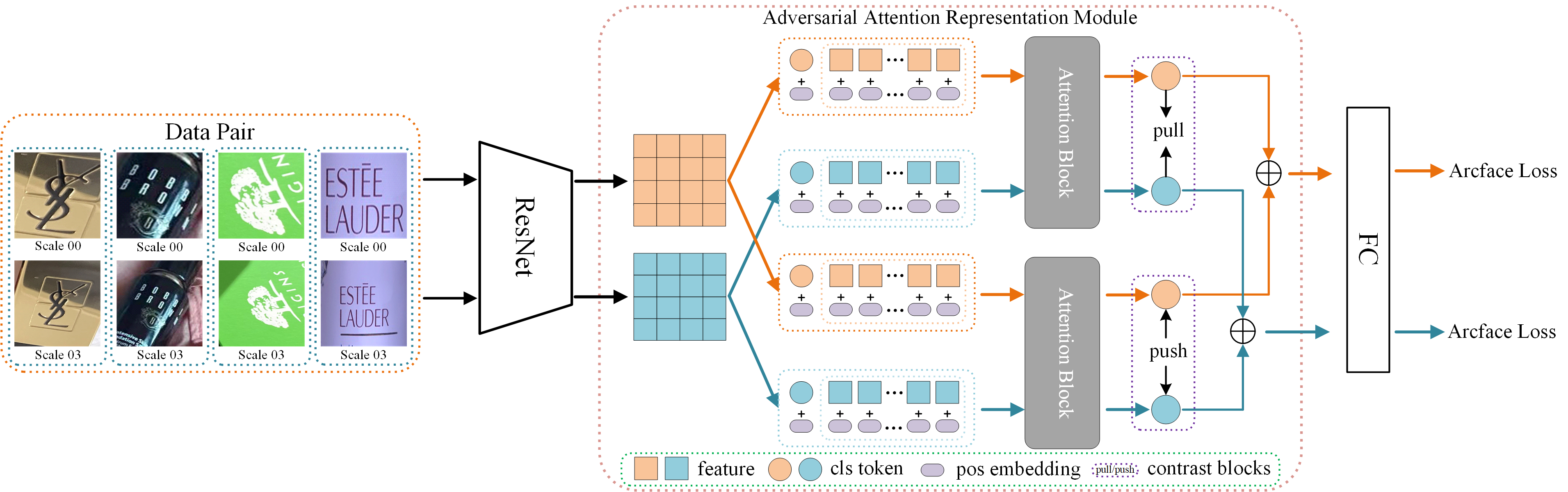}
\caption{\textbf{AAR architecture}; The AAR Module consists of two sub-module, each sub-module is composed of a attention block and a contrast block, the positive contrast pulls while the negative contrast pushes the two input embeddings. The input pairs are processed by the two sub-module to be adversarially attended on the common and uncommon features, respectively.}
\label{figure:5}
\end{figure*}

\subsection{Makeup216-Recg}
We also created a logo recognition dataset named \textbf{Makeup216-Recg}: the original bounding box was cropped as one branch, named as \textbf{Scale00}; and the original bounding box was enlarged by the scale of 0.3 and cropped as another branch, defined as \textbf{Scale03}. The training, validation and test sets of Makeup216-recg are not rearranged but inherited from Makeup216, and the detailed object number of training, validation and test sets is 23,046, 6,150, 7,822.

\subsection{Evaluation Procotol}
To investigate the effect of distractor samples on logo recognition, two logo recognition protocols were built. \textbf{Procotol 1-'w/o dis' } is defined as having no distractor samples, while \textbf{Procotol 2-'w dis'} is defined as having distractor samples. For Makeup216, the distractor samples are from clothes subset of Logodet-3k; we first cropped two scales of logo objects to generate \textbf{LogoRecg-3k-Clothes} as we did for Makeup216, and then put it together with the database of Makeup216-Recg. The reason to choose clothes subset is that its object number (37,504, after removing brands that are same with Makeup216) is close to Makeup216 (37,018). For LogoRecg-3k-Clothes, the distractor samples are from Makeup216-Recg.

\section{Methodology}

The proposed framework is illustrated in Fig. \ref{figure:5}. The above and below input images are sampled from Scale00 and Scale03 of Makeup216-Recg, respectively. The two inputs are first re-scaled to the same size and then encoded by a shared backbone network (ResNet18 without final pooling and full-connection layers). Afterwards, a dual-attention module is employed to adversarially attend on the common logo subject and uncommon marginal background separately; the two inputs of the positive attention sub-module are the two outputs of the shared encoder, so does the negative sub-module. Finally, the two streams of features after positive and negative contrasts of Scale00 are added and followed by a shared linear layer, as in the case of Scale03; both the representation of Scale00 and Scale03 after the linear layer are constrained by the Arcface loss.

\subsection{Data Pair Process}
Current self-supervised (\citeauthor{chen2021exploring} \citeyear{chen2021exploring}; \citeauthor{grill2020bootstrap} \citeyear{grill2020bootstrap}) generated two input images by randomly augmenting one image so that the model can learn features common to both views; whereas in our model, the input image pair are not just from the same logo object, but one is the original bounding box, the other is the enlarged one with more marginal background. The reason for generating in that way is that the pure logo is not feature-rich enough to avoid the false recall as we discussed in Fig. \ref{figure:1} while marginal background information would be helpful. The scale of enlarged branch is chosen as 0.3 which would be experimental proven in the ablation study section. The input image pair are first re-scaled to the same size \( (128 \times 128)\), the reason to choose this size is that the length of the encoded feature sequences would be \(16\), which is a trade-off between the computational requirement and the efficiency (\citeauthor{beltagy2020longformer}\citeyear{beltagy2020longformer}). The two input images are separately randomly augmented by the following dimensions: 1) horizontal and vertical flip, 2) crop and resize, 3) affine and perspective, 4) color jittering. 

\subsection{Adversarial Attention Representation Module}

\subsubsection{Attention Blocks}
Each input image is processed by the backbone to generate encoded feature map \(\mathbf{x}\in R^{H\times W\times C} \) , where \( (H, W)\) is the size of the 2D feature, \( C \) is the number of channels, \( H = 4\), \( W = 4\), \( C = 512\). The encoded feature map is equally divided into 16 patches and flatten into a sequence \(\mathbf{x}_p\in R^{16\times(1\times 1\times C)} \)  , each patch embedding represents features of a local region with dimension being \( 1\times1\times512\). A classification token \(\mathbf{x}_{cls}\in R^{ 1\times512}\)  is put at the left position of the patches embeddings (\(\mathbf{z}^0= \mathbf{x}_{cls}\)), thereby, the dimension of the sequence embeddings (\(\mathbf{z}\)) would be \( 17\times512\). Learnable position embedding \(\mathbf{x}_{pos}\) are also added to the patches embeddings to retain positional information as done by ViT (\citeauthor{dosovitskiy2020image} \citeyear{dosovitskiy2020image}). The sequence embeddings of one input image is constructed as:
\begin{equation}
\mathbf{z} = [\mathbf{x}^0_{cls}; \mathbf{x}^1_p ;\mathbf{x}^2_p;...;\mathbf{x}^{16}_p ] + \mathbf{x}_{pos}
\end{equation}
We are not only concerned with attention between patches of an input image, but also the attention between the two input images, so the sequence of Scale03 is appended to the sequence of Scale00, and the sequence of Scale00 can guide the sequence of Scale03 to find the attention area of logo subject. Therefore, the overall input sequence embeddings of two input images would be:
\begin{equation}
\mathbf{z}_{i} = [\mathbf{x}^0_{cls}; \mathbf{x}^1_p; ... ; \mathbf{x}^{16}_p; \mathbf{x}^{\prime17}_{cls}; \mathbf{x}^{\prime18}_p; ... ; \mathbf{x}^{\prime33}_p ] + \mathbf{x}_{pos}  
\end{equation}
where \(\mathbf{x}\) and \(\mathbf{x}^\prime\) indicate the feature map of Scale00 and Scale03, respectively;  \(\mathbf{x}^0_{cls}\) and  \(\mathbf{x}^{\prime17}_{cls}\) are the classification tokens of Scale00 and Scale03; \(\mathbf{x}_{pos}\)  is the position embeddings of the overall sequences with dimension of \(34\times512\).

The overall input sequence embeddings are then processed by two attention blocks, each attention block constitutes of two-depth transformer encoder (\citeauthor{dosovitskiy2020image} \citeyear{dosovitskiy2020image}; \citeauthor{vaswani2017attention} \citeyear{vaswani2017attention}) with 8 multi-head number. The output sequence embeddings of one attention block are:
\begin{equation}
\mathbf{z}_{o} = \mathcal{F} (\mathbf{z}_{i})
\end{equation}The output corresponding to the positions of the input classification tokens is served as the representations of Scale00 and Scale03, i.e. \(\mathbf{z}^0_{o1}\) and \(\mathbf{z}^{17}_{o1}\) . The other attention block produces \(\mathbf{z}^0_{o2}\) and \(\mathbf{z}^{17}_{o2}\). They are then processed by two contrast blocks.

\subsubsection{Contrast Blocks}
Two kinds of contrast blocks are designed, one is the positive contrast which pulls the two embeddings, and we minimize their negative cosine similarity:
\begin{equation}
\mathcal{D}_{pos}(\mathbf{z}^0_{o1},\mathbf{z}^{17}_{o1}) = - \frac{\mathbf{z}^0_{o1}\cdot\mathbf{z}^{17}_{o1}}{\left\|\mathbf{z}^0_{o1}\right\|\cdot\left\|\mathbf{z}^{17}_{o1}\right\|}
\end{equation}
the other is the negative contrast which pushes the two embeddings, and their positive consine similarity is minimized:

\begin{equation}
\mathcal{D}_{neg}(\mathbf{z}^0_{o2},\mathbf{z}^{17}_{o2}) = \frac{\mathbf{z}^0_{o2}\cdot\mathbf{z}^{17}_{o2}}{\left\|\mathbf{z}^0_{o2}\right\|\cdot\left\|\mathbf{z}^{17}_{o2}\right\|}
\end{equation}
The positive and negative contrast blocks are trained simultaneously to optimize the two similarities. The effect of the positive contrast block combined with the attention block is adversarial to the negative contrast block combined with the attention block, which is thus called an adversarial attention representation (AAR) module in this work.

The aim of the positive contrast is to learn the common features of the two input images, while the negative contrast is learning the uncommon features of the two inputs. AAR module is proved by the experiment below that it has two important advantages: 1) it can discriminate the logo subject and auxiliary marginal background; 2) it can improve the representation compared with the merely positive contrastive learning.

Each input image produces two adverse features after AAR module, which are added to obtain more comprehensive features, and then followed by a fully connected layer. Finally, both the final embedding of the two inputs are constrained by Arcface loss.  In the validation and test phases, only the final embedding of Scale03 are used to compare with that of the logos in the database.

\subsection{Loss Functions}
Two types of losses are exploited in the framework. The first is cosine similarity loss, which aims to compare the two classification token embedding after attention block. The positive contrast uses regular cosine similarity \(  f_{cos}\) while the negative contrast applies the adverse of cosine similarity \(  -f_{cos}\) . The total contrast loss function would be:

\begin{equation}
\mathcal{L}_{con} = \mathcal{D}_{pos(z_1, z_2)} +\lambda_{neg}  \mathcal{D}_{neg(z_3, z_4)}
\end{equation}
where \(z_1\) and \(z_2\) are the positive contrast embedding pairs, \(z_3\) and \(z_4\) are the negative contrast embedding pairs, \(\lambda_{neg}\) is the weight of the negative contrast loss. The other loss is metric loss, where we adopted Arcface loss \(\mathcal{L}_{arc}\) on each final embedding. The overall metric loss would be constructed as: 
\begin{equation}
\mathcal{L}_{metr} = \mathcal{L}_{arc(p_1)} + \mathcal{L}_{arc(p_2)}
\end{equation}
where \(p_1\) and \(p_2\) are final embeddings of two input images. By simply adding the contrast and metric loss, we get the training objective of AAR:
\begin{equation}
\mathcal{L}_{AAR} = \mathcal{L}_{metr} + \mathcal{L}_{con}
\end{equation}
An ablation about \(\lambda_{neg}\) is conducted in \textbf{\textit{Appendix B}}.

\section{Experiment}

\subsection{Implementation Details}

Experiments were carried out on two dataset: Makeup216-Recg and LogoRecg-3k-Clothes, and on two recognition tasks: close-set and open-set. Two protocols designed above were used to evaluate the performance. 

The framework was trained on one NVIDIA GeForce RTX 2080 Ti with batch size of 256; the learning rate was fixed at 1e-4, and Adam optimizer was adopted to optimize the whole framework. 

\begin{table}[h]
\centering

\resizebox{0.475\textwidth}{!}{
\begin{tabular}{cccccc}
\hline
 \multirow{2}*{Approaches}& \multirow{2}*{Scale}&\multicolumn{2}{|c}{Procotol 1}&\multicolumn{2}{|c}{Procotol 2}\\  
  & &@1&@5&@1&@5 \\
 \hline 
 Arcface (\citeauthor{wang2018additive} \citeyear{wang2018additive})&00&84.93 & 85.15 & 81.51 & 82.17	\\
 Triplet (\citeauthor{schroff2015facenet} \citeyear{schroff2015facenet})&00&82.67 & 83.13 & 74.55 & 75.68 \\
 CrossEntropy (\citeauthor{de2005tutorial} \citeyear{de2005tutorial})&00&87.89 & 88.15 & 81.73 & 82.93\\
 BYOL w/ Arc (\citeauthor{grill2020bootstrap} \citeyear{grill2020bootstrap})&00\&00& 87.05&87.50 & 83.24 & 84.04\\
 BYOL w/ Arc (\citeauthor{grill2020bootstrap} \citeyear{grill2020bootstrap})&00\&03& 85.24& 85.84& 80.68& 81.37 \\
 SimSiam w/ Arc (\citeauthor{chen2021exploring} \citeyear{chen2021exploring})&00\&00&87.11&87.82&82.39&83.11 \\
 SimSiam w/ Arc (\citeauthor{chen2021exploring} \citeyear{chen2021exploring})&00\&03&85.65&86.33&81.47&82.04\\
 \hline
  \textbf{AAR}&00\&03&\textbf{88.23}& \textbf{88.43} & \textbf{85.29} & \textbf{86.01}		\\
 
 \hline
\end{tabular}
}
\caption{Comparison results of different approaches on Makeup216-Recg; "@1" and "@5" indicate \%mAP@Top1 and \%mAP@Top5, respectively; "Arc" indicates Arcface.}
\label{table:2}
\end{table}

\subsection{Ablation study}
\subsubsection{AAR}
We compared AAR with several classic methods on Makeup216-Recg in a close-set way, and use retrieval mAP (\citeauthor{weyand2020google} \citeyear{weyand2020google}) to measure the performance, see Tab. \ref{table:2}. ResNet18 (\citeauthor{he2016deep} \citeyear{he2016deep}) is the backbone for all the methods, ResNet18 with Arcface is chosen as the baseline. Triplet and Arcface are compared for their effectiveness to learn discriminative metric embeddings. The simplest method is Classifier with CrossEntropy loss, where we train a linear classifier but extract the embedding before the classifier in the retrieval stage. SimSiam and BYOL which compare two augmented views of one image has some similarity with our constrat blocks are also considered, and it was appended by Arcface loss. The inputs for Classifier, Triplet and Arcface are only from Scale00 since they have no contrast modules, while the inputs for BYOL and SimSiam are two streams, where we compared Scale00\&03 and Scale00\&00. Semi-hard mining was also applied to all except Classifier. 

Triplet performs worse than baseline since it is very sensitive to the mining strategy (\citeauthor{wang2018additive} \citeyear{wang2018additive}). Classifier outperforms baseline suggesting that the Classifier may have been overfitted on the training set and therefore performs better on the close-set dataset. BYOL with Arcface at the scales of 00\&00 can enhance performance by about 2.2\% (protocol 1) and 1.8\% (protocol 2) compared with baseline, which indicates that the contrast play a considerable role in improving feature representation. We also trained BYOL at scales 00\&03 and used scale00 in the evaluation process. The declined performance for BYOL at scales of 00\&03 compared to that at 00\&00 indicates that marginal background poses some difficulties to learn representations. 

All the AAR related approaches are conducted at scales of 00\&03. AAR outperform baseline by the largest margin, approximate 3.3\% and 3.8\% on procotol 1 and procotol 2, respectively. The larger increase on protocol 2 compared with that on protocol 1 suggests that AAR is more robust in avoiding false recall when faced with distractor samples. 

\begin{table}[h]
\centering

\resizebox{0.475\textwidth}{!}{
\begin{tabular}{ccccccc}
\hline
 \multirow{2}*{Attn} & \multirow{2}*{Pos Con}& \multirow{2}*{Neg Con} & \multicolumn{2}{|c|}{Procotol 1}&\multicolumn{2}{|c}{Procotol 2}\\
  & & & @1 & @5 & @1 & @5 \\
 \hline
  & & & 84.57 & 84.8 & 80.22 & 80.81	\\
  & \checkmark & & 83.45 & 83.72 & 79.42 & 79.84	\\
  & & \checkmark& 83.37 & 83.73 & 79.83 & 80.47	\\
  & \checkmark& \checkmark & 84.00	& 84.27 & 79.44 &79.97 \\
  \checkmark& & &85.51 & 85.82 & 82.16 & 82.89 	\\
  \checkmark& \checkmark& & 88.04 & 88.15 & 83.78 & 84.61	\\
  \checkmark& &\checkmark&86.91 & 87.08 & 82.91 & 83.72	\\
  \checkmark&\checkmark&\checkmark&\textbf{88.23}& \textbf{88.43} & \textbf{85.29} & \textbf{86.01}\\
 \hline
\end{tabular}
}
\caption{Ablation results of AAR on Makeup216-Recg. "Attn" indicates attention block, "Pos Con" indicates positive contrast and "Neg Con" indicates negative contrast. }
\label{table:3}
\end{table}

An ablation study of AAR was also conducted (Tab. \ref{table:3}). In absence of attention, the performances are much reduced by at least about 3.6\% (protocol 1) and 5.1\% (protocol 2) compared with AAR, and even worse than the baseline, which suggests that marginal backgrounds can be intrusive if they are not taught to focus on the logo subject. Contrast does not help in the absence of attention, all positive, negative and adversarial contrast perform worse than no contrast, the reason is that it is quite difficult to learn common features of two different input images even they have some common features, the shared encoder gets confused if the two different inputs are forced to be the same. In the presence of attention, contrast improves a lot, positive contrast is better than negative contrast, reaching the best when positive and negative are combined (AAR). The positive contrast learns the common logo subjects and the negative contrast learns the uncommon marginal background as we expected, which, if added, are more comprehensive to represent logos. 

\subsubsection{Enlarged Scale}
\begin{figure}[h]
\centering
\includegraphics[width=0.48\textwidth ]{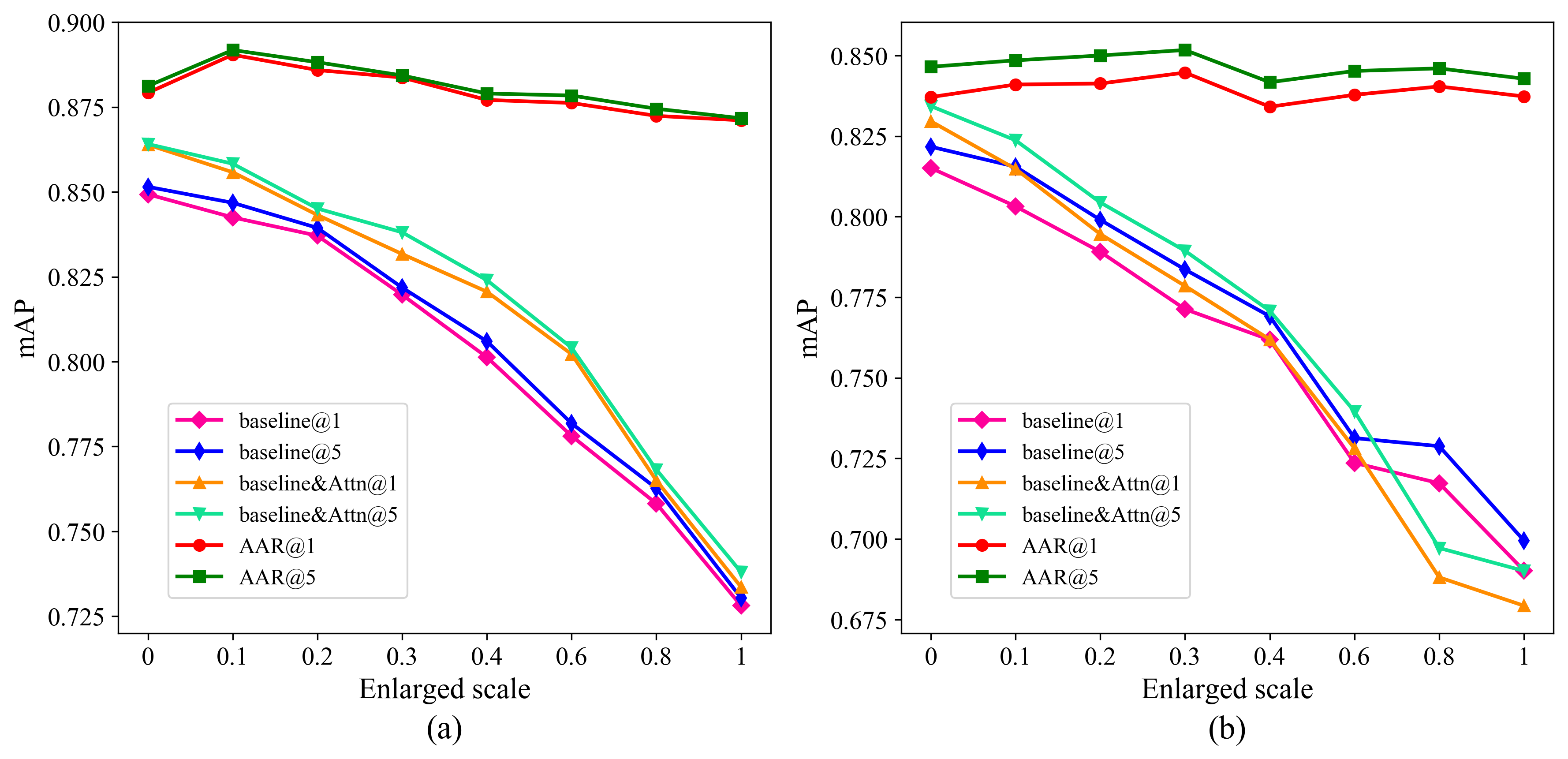}
\caption{Influence of enlarged scale on the performances, AAR is largely superior to baseline and baseline\&Attn, especially at larger scales; (a) protocol 1; (b) protocol 2.}
\label{figure:6}
\end{figure}

Various enlarged scales are selected as inputs to study how enlarged scale influences baseline and AAR, see Fig. \ref{figure:6}. It can be found that the improvement of AAR over baseline is significant on both protocols, their gap is not obvious at scale00 but baseline are gradually left behind by AAR as the scale increases. This means that too much marginal background introduces noise, and it is difficult for baseline to know where the logo subject is and where the marginal background is that should be payed less attention; whereas for AAR, it can distinguish between the two to suppress noise interference.

Further, they have completely different trends, the performance of the baseline keeps decreasing sharply with increasing scale, while AAR first increases gently and then decreases slightly, peaking at scale01 on protocol 1 and at scale03 on protocol 2. It suggests that marginal background is useful for AAR, and proper background can help it to obtain more comprehensive features while too much background would bring some disturbance. AAR still retains superiority at scale10 compared to that of the baseline at scale00.

In addition, the lift of AAR over the baseline is higher on protocol 2 than that on protocol 1. This is a very important contribution of AAR because in practical industrial applications, the data used for logo identification contains a large number of distractor samples. That is also why we choose the scale where it peaks on protocol 2 as our suggested scale.

Finally, we added a two-depth transformer encoder for baseline identical to AAR, called "baseline\&Attn" in Fig. \ref{figure:6}, and find that baseline with attention does not avoid decreasing with increasing scale. Therefore, it can be concluded that the advantage of AAR is not only due to the transformer encoder, but to the sophisticated design of the attention adversarial representation module.

We sampled several retrieval results to show the differences of AAR and baseline, see \textbf{\textit{Appendix C}}, i.e. the recalled images of AAR are similar to query images at a broader context while baseline only concerned with logo subject.

\subsubsection{Adversarial contrast}
\begin{table}[h]
\centering
\resizebox{0.475\textwidth}{!}{
\begin{tabular}{cccccc}
\hline
 \multirow{2}*{Approaches} & \multirow{2}*{Scales} & \multicolumn{2}{|c|}{Procotol 1}&\multicolumn{2}{|c}{Procotol 2}\\
  & & @1 & @5 & @1 & @5 \\
 \hline
 AAR w/ Attn, Pos Con & \multirow{2}*{00\&03} & 88.04 & 88.15 &	83.78 &	84.61 \\
 \textbf{AAR} & & \textbf{88.23}& \textbf{88.43} & \textbf{85.29} & \textbf{86.01} \\
 AAR w/ Attn, Pos Con & \multirow{2}*{00\&00} & 86.49 &	86.62 &	82.91 &	83.76 \\
 \textbf{AAR} & & \textbf{87.92}& \textbf{88.12} & \textbf{83.71} & \textbf{84.65} \\
 \hline
\end{tabular}
}
\caption{The advancement of adversarial contrast over single positive contrast.}
\label{table:4}
\end{table}
  
We would like to illustrate the superiority of adversarial contrast over single positive contrast see Tab. \ref{table:4}, and hope this idea would bring some inspiration to people. It can be readily understood that the performance is improved at scale 00\&03 because AAR can learn both logo subject and auxiliary marginal background to enhance the logo features. More excitingly, the adversarial contrast still outperforms the single positive contrast at scales 00\&00 where the two inputs are from the same image. This is evidence that adversarial contrast leads to improved representations compared to single positive contrasts.

\subsection{Open set }
\subsubsection{Open-set dataset}
Open set logo recognition means that the logo classes in the test set do not appear in the training process, so we randomly selected 40\% of the logo classes from Makeup-Recg as the test set. The remaining 60\% logo classes were divided into training and validation sets in a close-set way with a ratio of 8:2, and also by the order of logo products instead of randomly.

\subsubsection{Results}

\begin{table}[h]
\centering
\resizebox{0.475\textwidth}{!}{
\begin{tabular}{cccccc}
\hline
 \multirow{2}*{Approaches}& \multirow{2}*{Scale}&\multicolumn{2}{|c|}{Procotol 1}&\multicolumn{2}{|c}{Procotol 2}\\
 & &@1&@5&@1&@5 \\
 \hline 
 Arcface  &00& 71.99&74.12&59.24&61.34	\\
 Triplet&00&61.31&64.59&46.97&49.28 \\
 CrossEntropy&00&71.07&73.68&55.74&58.68\\
 BYOL /w Arc &00\&00&74.38&76.05&61.84&63.57 \\
 BYOL /w Arc &00\&03&69.92&72.15&56.72&58.87 \\
 SimSiam /w Arc &00\&00&70.71&72.90&55.99&58.07 \\
 SimSiam /w Arc &00\&03&69.08&71.76&56.76&58.65\\
 \hline
 \textbf{AAR} &00\&03&\textbf{75.17}&\textbf{77.20}&\textbf{64.79}&\textbf{66.58}		\\
 \hline
\end{tabular}
}
\caption{Open-set comparison on Makeup216-Recg}
\label{table:5}
\end{table}

It can be seen from Tab. \ref{table:5} that Triplet is the worst; Arcface (baseline) is better than CrossEntropy which is opposite with the case of close-set, confirming the advantage of Arcface in the case of open-set. It does not increase too much at scales of 00\&00 and even get declined than baseline at scales of 00\&03 for BYOL, and it perform worse than baseline at both scale pairs for SimSiam, suggesting that these two methods which were first introduced into feature learning for large-scale image datasets are not suitable for learning logo features. In contrast, AAR promotes the best by 3.1\% and 5.4\% on protocol 1 and protocol 2, respectively; and it improves more on protocol 2 compared to the case of close set. This confirms the competitiveness of AAR in solving open-set logo recognition problems.

\begin{table}[h]
\centering

\resizebox{0.475\textwidth}{!}{
\begin{tabular}{ccccccc}
\hline
 &\multirow{2}*{Approaches}&\multirow{2}*{Scale}&\multicolumn{2}{|c|}{Procotol 1}&\multicolumn{2}{|c}{Procotol 2}\\  
  & & &@1&@5&@1&@5 \\
 \hline 
 
  \multirow{8}*{Close-set}&Arcface&00&86.7&88.33&85.61&87.15	\\
  &Triplet&00&86.99&88.58&85.85&87.58\\
 &CrossEntropy&00&\textbf{87.7}&\textbf{89.38}&86.36&88.05\\
 &BYOL /w Arc&00\&00&86.93&88.51&85.74&87.37\\
 &BYOL /w Arc &00\&03&85.67&87.25&84.77&86.33 \\
 &SimSiam /w Arc &00\&00&87.02&88.73&86.07&87.73 \\
 &SimSiam /w Arc &00\&03&86.02&87.71&85.09&86.73 \\
 &\textbf{AAR} &00\&03&87.5&89.02&\textbf{87.01}&\textbf{88.62}		\\
 \hline
 
 \multirow{8}*{Open-set}&Arcface&00&76.75&78.02&71.29&72.36 \\
 &Triplet&00&71.07&72.78&63.26&65.12 \\
 &CrossEntropy&00&76.54&77.86&70.35&71.43 \\
 &BYOL /w Arc& 00\&00 &76.75&78.05&72.02&73.2 \\
 &BYOL /w Arc& 00\&03&71.55&73.12&65.82&67.52 \\
 &SimSiam /w Arc &00\&00&77.05&78.19&72.23&73.4  \\
 &SimSiam /w Arc &00\&03&72.59&74.21&66.14&67.96 \\
&\textbf{AAR}&00\&03&\textbf{78.07}&\textbf{79.45}&\textbf{73.7}&\textbf{74.84}		\\
 \hline
\end{tabular}
}
\caption{Close-set and open-set comparisons on LogoRecg-3k-Clothes}
\label{table:6}
\end{table}

\subsection{LogoRecg-3k-Clothes}

We chose LogoRecg-3k-Clothes to verify the generalizability of AAR (Tab. \ref{table:6}). The training, validation, and testing sets are split in the same way as Makeup216 both for the cases of close-set and open-set, except that they are all randomized. In the case of close-set, stronger feature learning methods like BYOL and Simsiam do not make a difference while Classifier wins by a very small margin, which may caused by the simplicity of LogoRecg-3k-Clothes. However, AAR still performs best on protocol 2 showing the robustness facing distractor samples. 
As for open-set, the performance of different methods is similar with that on Makeup216: Arcface is better than Classifier and Triplet; BYOL and SimSiam are better than Arcface but with smaller lift on scales of 00\&00, and the performances are greatly reduced on scales of 00\&03 when marginal background are introduced. Still, AAR outperforms the baseline on both protocols, by 1.4\%, 2.4\%, respectively. 


\subsection{Visualization}

\begin{figure}[h]
\centering
\includegraphics[width=0.48\textwidth ]{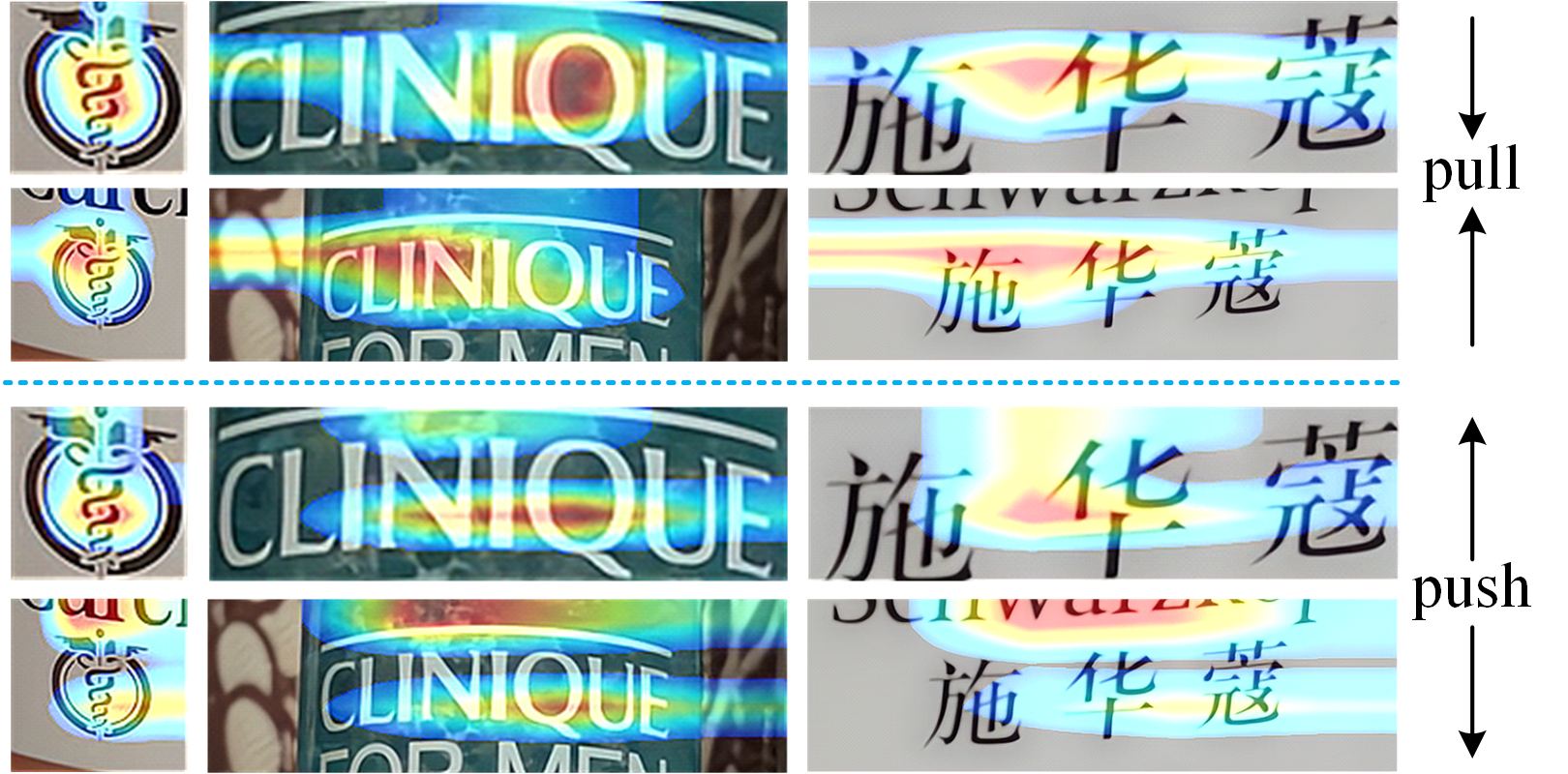}
\caption{Visual explanations of adversarial contrasts: pull contrast makes attention on logo subjects while push contrast makes attention on uncommon marginal background. }
\label{figure:7}
\end{figure}

In order to visually explain how the AAR module works, GradCAM (\citeauthor{selvaraju2017grad}\citeyear{selvaraju2017grad}) was employed to plot the attentional location heatmap on the test set of Makeup216. The gradient is obtained by computing the derivative of the best matching score with respect to the activations of the attention layer, the gradient is global-average-pooled as a weight and then combined with the forward activation feature map to generate the final heatmap, as shown in Fig. \ref{figure:7}. Basically, the heatmaps conform to the aim of AAR: for positive contrast, both scales of images are focused on the common logo subjects; whereas for negative contrast, scale00 are attended more on the logo subjects while scale03 are more on the marginal background. It reveals that AAR can discriminate the common and uncommon regions, and if the two are combined, more comprehensive feature representation can be acquired.


\section{Conclusion}
In this work, we mainly focused on logo recognition and introduced a logo dataset, Makeup216, which shows the great challenge of practical logo recognition. It was found that logos could be easily falsely identified due to the lack of rich features. Therefore, we supposed that the features of the logo can be complemented by the marginal background, and thus proposed an adversarial attention representation (AAR) module to take full advantage of the marginal background as auxiliary information. Experiments proved that the AAR module could discriminate the common logo subject and uncommon marginal background, and better representation and performance can be achieved if the two are combined.

\bibliographystyle{aaai}
\bibliography{ref.bib}

\end{document}